\pdfoutput=1

\documentclass[11pt]{article}

\usepackage{format/acl}

\usepackage{times}
\usepackage{latexsym}

\usepackage[T1]{fontenc}

\usepackage[utf8]{inputenc}

\newcommand{\stitle}[1]{\vspace{1ex} \noindent{\bf #1.}}

\usepackage{microtype}
\usepackage{booktabs}
\usepackage{multicol}
\usepackage{multirow}
\usepackage{graphics}
\usepackage{tikz}
\usepackage{xspace}
\usepackage{amssymb}
\usepackage{enumitem}
\usepackage{tcolorbox}
\usepackage{amsmath}
\usepackage{float}

\usepackage{cleveref}
\crefformat{section}{\S#2#1#3}
\crefformat{subsection}{\S#2#1#3}
\crefformat{subsubsection}{\S#2#1#3}
\crefrangeformat{section}{\S#3#1#4 to~\S#5#2#6}
\crefmultiformat{section}{\S#2#1#3}{ and~\S#2#1#3}{, #2#1#3}{ and~#2#1#3}
\usepackage{refstyle}
\Crefformat{figure}{#2Fig.~#1#3}
\Crefmultiformat{figure}{Figs.~#2#1#3}{ and~#2#1#3}{, #2#1#3}{ and~#2#1#3}
\Crefformat{table}{#2Tab.~#1#3}
\Crefmultiformat{table}{Tabs.~#2#1#3}{ and~#2#1#3}{, #2#1#3}{ and~#2#1#3}
\Crefformat{appendix}{Appx.~\S#2#1#3}
\crefformat{algorithm}{Alg.~#2#1#3}

\makeatletter
\def\hlinewd#1{%
\noalign{\ifnum0=`}\fi\hrule \@height #1 %
\futurelet\reserved@a\@xhline}
\makeatother

\newcommand{\model}{\mbox{\textsc{Lattice}}\xspace}
\newcommand{\mojang}{\includegraphics[height=1em]{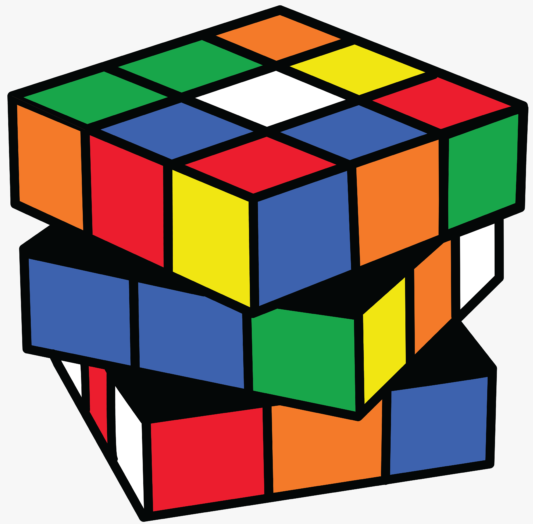}}

\title{Robust (Controlled) Table-to-Text Generation with\\ %
Structure-Aware Equivariance Learning
}

\author{Fei Wang,\;~Zhewei Xu,\;~Pedro Szekely \and Muhao Chen \\
  Department of Computer Science \& Information Sciences Institute\\
  University of Southern California \\
  \texttt{\{fwang598,zheweixu,szekely,muhaoche\}@usc.edu} 
  }

\begin{document}
\maketitle

\begin{abstract}

Controlled table-to-text generation seeks to generate natural language descriptions for highlighted subparts of a table. Previous SOTA systems still employ a sequence-to-sequence generation method, which merely captures the table as a linear structure and is brittle when table layouts change. 
We seek to go beyond this paradigm by (1) effectively expressing the relations of content pieces in the table, and (2) making our model robust to content-invariant structural transformations. 
Accordingly, we propose an equivariance learning framework, \model~(\mojang), which encodes tables with a structure-aware self-attention mechanism. 
This prunes the full self-attention structure into an order-invariant graph attention that captures the connected graph structure of cells belonging to the same row or column, and it differentiates between relevant cells and irrelevant cells from the structural perspective. 
Our framework also modifies the positional encoding mechanism to preserve the relative position of tokens in the same cell but enforce position invariance among different cells. 
Our technology is free to be plugged into existing table-to-text generation models, and has improved T5-based models to offer better performance on ToTTo and HiTab. 
Moreover, on a harder version of ToTTo, we preserve promising performance, while previous SOTA systems, even with transformation-based data augmentation, have seen significant performance drops.\footnote{Our code is available at \url{https://github.com/luka-group/Lattice}.}
\end{abstract}
\section{Introduction}

Table-to-text generation seeks to generate natural language descriptions for content and entailed conclusions %
in tables.
It is an important task that not only makes ubiquitous tabular data more discoverable and accessible,
but also supports downstream tasks of tabular semantic retrieval \cite{wang2021retrieving},
reasoning \cite{gupta2020infotabs}, fact checking \cite{chen2019tabfact,wang2021table} and table-assisted question answering \cite{chen2020hybridqa}.
While rich and diverse facts can be presented in a table,
the controlled table-to-text generation task, which generates focused textual descriptions for highlighted subparts of a table, has garnered much attention recently \citep{parikh2020totto, kale2020text, cheng2021hitab}.

\begin{figure}[t]
  \begin{center}
    \includegraphics[width=7.5cm]{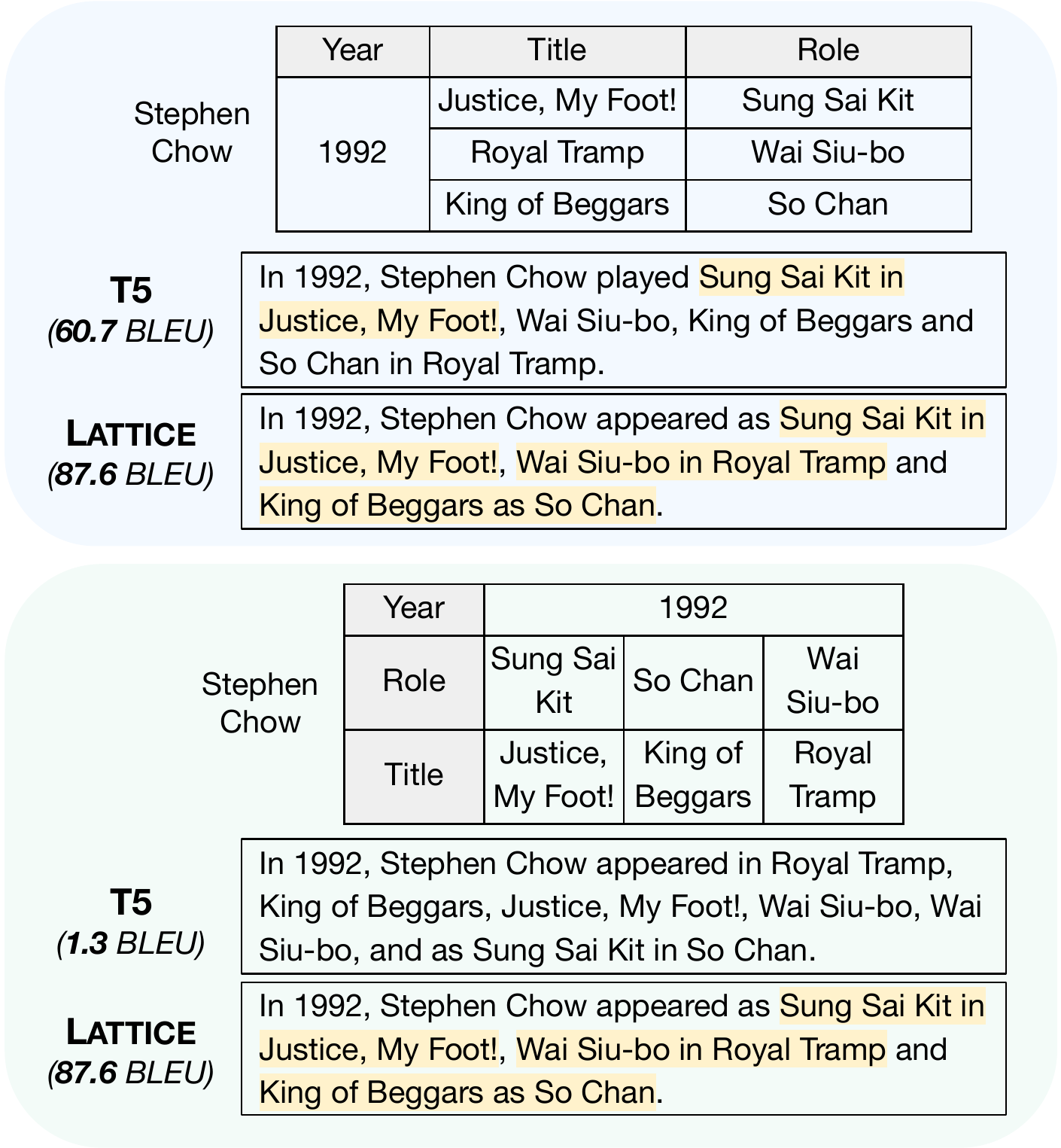} %
  \end{center}
  \caption{Description generation on content-equivalent tables with different layouts by T5 and \model\footnotemark. %
  Correct film-role pairs in generations are in orange. We report also the BLEU-4 score of each generation. T5 is brittle to layout changes, while \model returns consistent results. 
  }
  \label{fig/example}
\end{figure}
\footnotetext{{This example is from the ToTTo dataset. Original film names and role names are too long. For presentation, we replace the actor name, film names and role names.}}

Prior studies on controlled table-to-text generation often employ a sequence-to-sequence generation method, which merely captures the table as a linear structure \citep{parikh2020totto, kale2020text, su2021plan}.
However, table layouts, though overlooked by prior studies,  %
are key to the generation from two perspectives. %
First, table layouts indicate the relations among cells that collectively present a fact, %
which are however not simply captured by a linearized table. %
For example, if we linearize the first table row-wise in \Cref{fig/example}, \textit{Wai Siu-bo} will be next to both \textit{Royal Tramp} and \textit{King of Beggers}, so that it is not clear this role belongs to which film.
Second, the same content can be equivalently expressed in tables with different layouts. While linearization simplifies the layout representation, %
it causes brittle generation when table layouts change. 
\Cref{fig/example} shows two tables with the same content but different layouts, %
for which the generations by T5 are largely inconsistent.

In this paper, we focus on improving controlled table-to-text generation systems by incorporating two properties: \emph{structure-awareness} and \emph{transformation-invariance}.
Structure-awareness, which seeks to understand cell relations indicated by the table structure, is essential for capturing contextualized cell information. 
Transformation-invariance, which seeks to make the model insensitive to content-invariant structural transformations (including transpose, row shuffle and column shuffle), is essential for model robustness.
However, incorporating structure-awareness and transformation-invariance into existing generative neural networks is nontrivial, especially when preserving the generation ability of pretrained models as much as possible.

We enforce the awareness of table layouts and robustness to content-invariant structural transformations on pretrained generative models with %
an equivariance learning framework, namely \textsc{\underline{L}}ayout \textsc{\underline{A}}ware and \textsc{\underline{T}}ransforma\textsc{\underline{T}}ion \textsc{\underline{I}}nvariant \textsc{\underline{C}}ontrolled Table-to-Text G\textsc{\underline{E}}neration (\model~\mojang).
\model encodes tables with a transformation-invariant graph masking technology. 
This prunes the full self-attention structure into an order-invariant graph-based attention that captures the connected graph of cells belonging to the same row or column, and differentiates between relevant cells and irrelevant cells from the structural perspective. 
\model also modifies the positional encoding mechanism to preserve the relative position of tokens within the same cell but enforces position invariance among different cells.
Our technology is free to be plugged into existing table-to-text generation models, and has improved T5-based models \citep{raffel2020exploring} %
on ToTTo \citep{parikh2020totto} and HiTab \citep{cheng2021hitab}. Moreover, on a harder version of ToTTo, we preserve promising performance, while previous SOTA systems, even with transformation-based data augmentation, have seen significant performance drops.

Our contributions are three-fold.
First, we propose two essential properties of a precise and robust controlled table-to-text generation system, i.e. structure-awareness and transformation-invariance.
Second, we %
demonstrate how our transformation-invariant graph masking technology can enforce these two properties, and effectively enhance a representative group of Transformer-based generative models, i.e. T5-based models, for more generalizable and accurate generation.
Third, in addition to experiments on ToTTo and HiTab benchmarks, we evaluate our model on %
a harder version of ToTTo with a special focus on robustness to content-invariant structural transformations.

\section{Method}
In this section, we first describe the preliminaries of content-invariant table transformations, base models and the input format for controlled table-to-text generation (\Cref{sec/method/base}). 
Then we introduce the technical details about how the transformation-invariant graph masking technology in \model enforces the model to be structure-aware and transformation-invariant (\Cref{sec/method/mask}). 
Finally, we present two alternative techniques for strengthening the transformation-invariance to be compared with \model (\Cref{sec/method/aug}). 

\subsection{Preliminaries}
\label{sec/method/base}

\begin{figure*}[t]
  \begin{center}
    \includegraphics[width=0.98\textwidth]{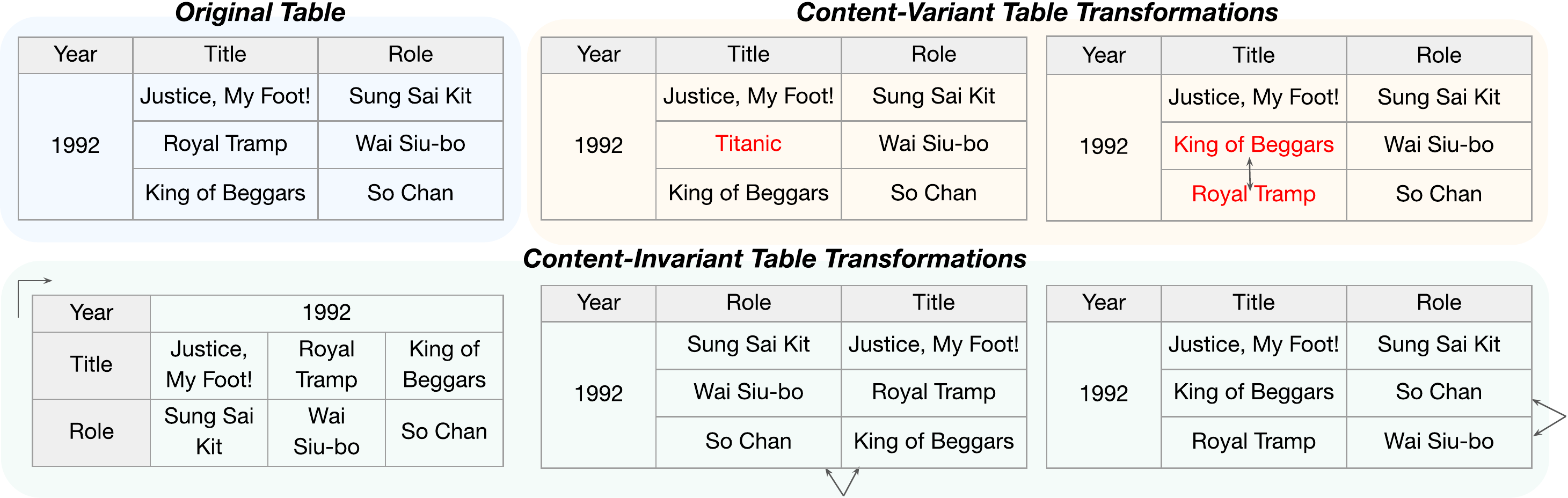}
  \end{center}
  \vspace{-0.5em}
  \caption{Examples of different types of table transformations. Arrows indicate how specific operations change the positions of tables components. Modifications causing semantic changes are in red. }
  \label{fig/transformation}
\end{figure*}

\stitle{Content-Invariant Table Transformations}
Tables organize and present information by rows and columns.
A piece of information is presented in a cell (with headers), which is the basic unit of a table.
Rows and columns are high-level units indicating relations among cells, and are combined to express more comprehensive information.
We discuss two categories of transformations that may be made on a table, as shown in \Cref{fig/transformation}.
First, \emph{content-variant} transformations modify or exchange a part of cells in different rows or columns, therefore changing the semantics of the table. 
In such cases, new tabular content are created to express information being inconsistent with the original table.
Second, %
\emph{content-invariant} transformations consist of operations that do not influence content within (combinations of) the same row or column, resulting in semantically equal (sub-)tables.
Specifically, such operations include transpose, row %
shuffle and column shuffle.
By performing any or a combination of such operations, we can present the same information in different table layouts.

\stitle{Base Models}
Pretrained Transformer-based generative models achieve SOTA performance on various text generation tasks \citep{raffel2020exploring,lewis2020bart}.
In order to adapt this kind of models to table-to-text generation, prior works propose to linearize the table into a textual sequence \citep{kale2020text,chen2020logical,su2021plan}.
Our method \model is model-agnostic and can be incorporated into any such models.
Following \citet{kale2020text}, we choose a family of the best performing models, T5 \citep{raffel2020exploring}, as our base models.
Models of this family are jointly pretrained on a series of supervised and self-supervised text-to-text tasks. %
Models can switch between different tasks by prepending a task-specific prefix to the input.
Our experiments (\Cref{sec/exp/robust,sec/exp/ablation}) point out that base models are brittle to content-invariant table transformations and can only capture limited layout information.

\stitle{Input Format}
Prior works \citep{kale2020text,chen2020logical,su2021plan} linearize (highlighted) table cells based on row and column indexes.
The input sequence often starts with the metadata of a table, such as page title and section title. 
Then, it traverses the table row-wise from the top-left cell to the bottom-right cell.
Headers of each cell can be either treated as individual cells or appended to the cell content.
Each metadata/cell/header field is separated with special tokens.
This linearization process suits the input to text-to-text generation models, yet discards much of the structural information of a table (e.g., two cells in the same column can be separated by irrelevant cells in the sequence, while the last cell and first cell in adjacent rows can be adjacent although they are irrelevant), and is sensitive to content-invariant table transformations.

\begin{figure*}[t]
  \begin{center}
    \includegraphics[width=0.98\textwidth]{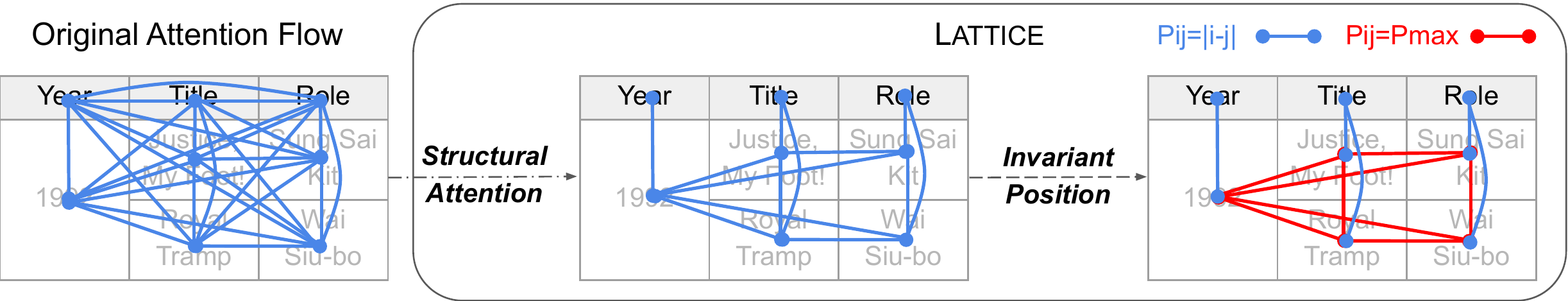}
  \end{center}
  \vspace{-0.5em}
  \caption{Attention flows of the base model and \model. 
  In this example, we adopt the input format which appends headers to each cell, so headers can be seen as part of the cell content. 
  We omit the attention flows among tokens within a cell, as they are in the same type of the flows between headers and corresponding cells.
  $P_{ij}$ represents the relative position between tokens at both ends of the attention flow, where $i$ and $j$ are absolute positions of tokens in the linearized table and $P_{max}$ is the max relative position allowed.
  The base model has a complete attention graph among all cells with relative positions based on linear distance. 
  \model prunes the attention flow based on the table layout and assigns transformation-invariant relative positions between cells.
  }
  \label{fig/lattice}
\end{figure*}

\subsection{Transformation-Invariant Graph Masking}
\label{sec/method/mask}

\model realizes equivariance learning by modifying the Transformer encoder architecture. It also improves the base model's ability of capturing structures of highlighted tabular content.
Specifically, we incorporate a structure-aware self-attention mechanism and a transformation invariant positional encoding mechanism in the base model
The workflow is shown in \Cref{fig/lattice}.

\stitle{Structure-Aware Self-Attention}
Transformer \cite{vaswani2017attention} adopts self-attention to aggregate information from all the tokens in the input sequence.
The attention flows form a complete graph connecting each token.
This mechanism works well for modeling sequences %
but falls short of capturing %
tabular structures. %
The non-linear layout structure reflects semantic relations among cells, hence should be captured by self-attention.

We incorporate structural information by pruning the attention flows. 
According to the %
nature of information arrangement in a table, two cells in neither the same row nor the same column are not semantically related, or at least the combination of them do not directly express information this table seeks to convey.
Intuitively, representations of these cells should not directly pass information to each other.
In \model, attention flows among tokens of structurally unrelated cells are removed from the attention graph, while those within the metadata, within each cell, and between metadata and each cell are preserved.
In this way, we also ensure the transformation-invariance property of the self-attention mechanism, since related cells in the same row or same column are all linked %
in an unordered way in the attention graph.
It is easy to show that for any individual cell, the links in the attention graph will remain the same after any content-invariant operations (\Cref{sec/method/base}) are applied.

\stitle{Transformation-Invariant Positional Encoding}
When calculating the attention scores between each pair of tokens, the base model captures their relative position in the sequence of linearized table as an influential feature.
Specifically, %
the attention flow from the $i$-th token to the $j$-th token is paired with a relative position $P_{ij} = |i - j|$.
This easily causes positional biases among distinct cells, since the relative positions in the sequence do not fully reflect relations among cells in the table.
Moreover, the relative position between the same token pair will change as the table layout change, which is the source of inconsistent generation shown in \Cref{fig/example}.

As discussed in \Cref{sec/method/base}, for a given cell, its relations with other cells in the same row or column should be %
equally considered.
It is natural to assign the same relative positions among (tokens of) cells in the same row or column, no matter how far their distance is in the linear sequence.
Meanwhile, we preserve the relative positions of tokens inside the same cell (or the metadata). 
Specifically, %
the relative position between the $i$-th token and the $j$-th token in the input sequence is
\begin{equation*}  
P_{ij} = P_{ji} = \left \{  
     \begin{array}{ll}
     |i-j|, & \text{if in the same field};   \\  
     P_{max}, & \text{otherwise};    
     \end{array}  
\right.  
\end{equation*} 
where ``same field'' means the two tokens are from the same cell or both of them are from the metadata, and $P_{max}$ is the max relative position allowed.
As a result, %
\model represents cells (and the metadata) in a way that is invariant to their relative positions in the sequence.
As content-invariant table transformations do not change the relations among cells in the table (i.e. whether two cells are from the same row or column), this positional encoding mechanism is transformation-invariant. 

\stitle{Training and Inference}
After obtaining the structure-aware and transformation-invariant table representation, \model conducts similar training and inference as the base model.
Given the linearized table $T_i$, its layout structure $S_i$,  and target sentence $Y_i = \{y^i_1, y^i_2, ..., y^i_{n_i}\}$, training minimizes the negative log-likelihood. For a dataset (or batch) with $N$ samples, the loss function is
$$L = - \frac{1}{N} \sum_{i=1}^{N} \sum_{j=1}^{n_i} \log P(y^i_j | y^i_{<j}, T_i, S_i).$$
During inference, the model generates a sentence token by token, where each time it outputs a distribution over a vocabulary. 

\subsection{Alternative Techniques}
\label{sec/method/aug}

In addition to the equivariance learning realized by tranformation-invariant graph masking, we present and compare with two alternative techniques.

\stitle{Layout-Agnostic Input}
The first technique is to adjust input sequences to be invariant to content-invariant table transformations.
A simple way is to reorder headers and cells by an arbitrary order not based on table layouts (e.g., lexicographic order) to form a sequence.
Special tokens to separate cells and headers should also include no layout information\footnote{For example, using \textit{<header>} instead of \textit{<row\_header>} and \textit{<column\_header>}.}.
As a result, this input format loses all information about table layouts to ensure transformation-invariance.

\stitle{Data Augmentation}
The second technique is data augmentation by content-invariant table transformation. 
This technique augment tables with different layouts to training data, seeking to enhance the robustness of the base model by exposing it to more diverse training instances.

Our experiments systematically compares these two techniques with tranformation-invariant graph masking in \Cref{sec/exp/robust}, revealing how directly performing equivariance learning from the perspective of neural network structure leads to better performance and robustness than using layout-agnostic input or data augmentation.

\section{Experiments}
In this section, we conduct experiments on two benchmark datasets. 
First, we introduce the details of datasets, baselines, evaluation metrics and our implementation (\cref{sec/exp/setting}).
Then, we show the overall performance of \model (\cref{sec/exp/main}).
After that, we analyze the model robustness on a harder version of the ToTTo dataset where content-invariant perturbations are introduced (\cref{sec/exp/robust}).
Finally, we provide ablation study on components of transformation-invariant graph masking (\cref{sec/exp/ablation}).

\subsection{Experimental Settings}
\label{sec/exp/setting}

\stitle{Datasets}
We evaluate our model on ToTTo \citep{parikh2020totto} and HiTab \citep{cheng2021hitab} benchmarks.
Details of them are described as follows: 

\begin{itemize}[leftmargin=*]
\setlength\itemsep{-0.1em}
\item \textbf{ToTTo:} 
An English dataset released under the Apache License v2.0. The dataset is dedicated to controlled table-to-text generation.
It consists of 83,141 Wikipedia tables, 120,761/7,700/7,700 sentences (i.e. descriptions of tabular data) for train/dev/test. 
Target sentences in test set are not publicly available. 
Each sentence is paired with a set of highlighted cells in a table, and each table has metadata including its page title and section title.
The dev and test sets can be further split into 2 subsets, i.e. overlap and non-overlap, according to whether the table exists in the training set.
\item \textbf{HiTab:}
An English dataset released under Microsoft's Computational Use of Data Agreement (C-UDA). It is intended for both controlled table-to-text generation and table-based question answering with a special focus on hierarchical tables.
It contains 3,597 tables, including tables from statistical reports and Wikipedia, forming 10,686 samples distributed across train (70\%), dev (15\%), and test (15\%).
Each sample consists of a target sentence and a table with highlighted cells and hierarchical headers.
\end{itemize}

\stitle{Evaluation Metrics}
We adopt three widely used evaluation metrics for text generation.
BLEU \citep{papineni2002bleu} is one of the most common metric for text generation based on $n$-gram co-occurrence. We use the commonly used BLEU-4 following prior works \citep{parikh2020totto, cheng2021hitab}.
PARENT \citep{dhingra2019handling} is a metric for data-to-text evaluation taking both references and tables into account.
BLEURT \citep{sellam2020bleurt} is a learned evaluation metric for text generation based on BERT \cite{devlin2019bert}.
Following prior studies \citep{parikh2020totto, cheng2021hitab}, we report all three metrics on ToTTo and the first two metrics on HiTab using the evaluation tool released by \citet{parikh2020totto}.

\stitle{Baselines}
We present baseline results of the following representative methods:

\begin{itemize}[leftmargin=*]
\setlength\itemsep{-0.1em}
\item \textbf{Pointer-Generator} \citep{gehrmann2018end}: An LSTM-based encoder-decoder model with attention and copy mechanism, first proposed by \citet{see2017get} for text summarization.
\item \textbf{BERT-to-BERT} \citep{rothe2020leveraging}: A Transformer-based encoder-decoder model, where the encoder and decoder are initialized with BERT \citep{devlin2019bert}.
\item \textbf{T5} \citep{kale2020text}: A pretrained generation model first proposed by \citet{raffel2020exploring}. The model is Transformer-based, pretrained on text-to-text tasks, and finetuned on linearized tables to offer the previous SOTA performance.
\end{itemize}

All the baseline results on ToTTo can be found in the official leaderboard\footnote{\url{https://github.com/google-research-datasets/ToTTo}}, except for T5-small and T5-base, for which we reproduce the results on dev set reported by \citet{kale2020text} and submit the predictions on hidden test set to the leaderboard.
For HiTab, we run T5 and \model using our replication of the linearization process introduced by \citet{cheng2021hitab}\footnote{According to the authors, their linearization process needs unreleased raw excel files. We reproduce it with released tables which results in less precise and informative inputs.}. Results of other baselines are from \citet{cheng2021hitab}.

\stitle{Implementation Details}
We adopt the pretrained model weights released by \citet{raffel2020exploring}.
Specifically, we use T5-small and T5-base\footnote{Although a previous study \citep{kale2020text} has obtained better results using the much larger T5-3B, we were not able to run that model on our equipment even with a batch size of 1 due to the overly excessive GPU memory usage.}.
For finetuning, we use a batch size of 8 and a constant learning rate of $2e^{-4}$.
Following \citet{kale2020text} and \citet{cheng2021hitab}, all input sequences are truncated to a length of 512 to accommodate the limit of the pretrained models.
\model does not add any parameters to the base model, so \model (T5-small) has 60 million parameters and \model (T5-base) has 220 million parameters, same as the base models.
For ToTTo, we use a beam size of 4 to generate sentences with at most 128 tokens.
For HiTab, we use a beam size of 5 to generate sentences with at most 60 tokens following \citet{cheng2021hitab}.
Our implementation is based on Pytorch \citep{paszke2019pytorch} and Transformers \citep{wolf2020transformers}.
We run experiments on a commodity server with a GeForce RTX 2080 GPU.
It takes about 0.5 hour to train \model (T5-small) for 10,000 steps and about 1 hour to train \model (T5-base) for 10,000 steps.
Considering different sizes of two datasets, we train models for 150,000 steps on ToTTo, and for 20,000 steps on HiTab.
Results of \model on ToTTo dev set and HiTab are average of multiple runs.
For ToTTo test set, we report the results on official leaderboard.

\begin{figure}[t]
  \begin{center}
    \includegraphics[width=\columnwidth]{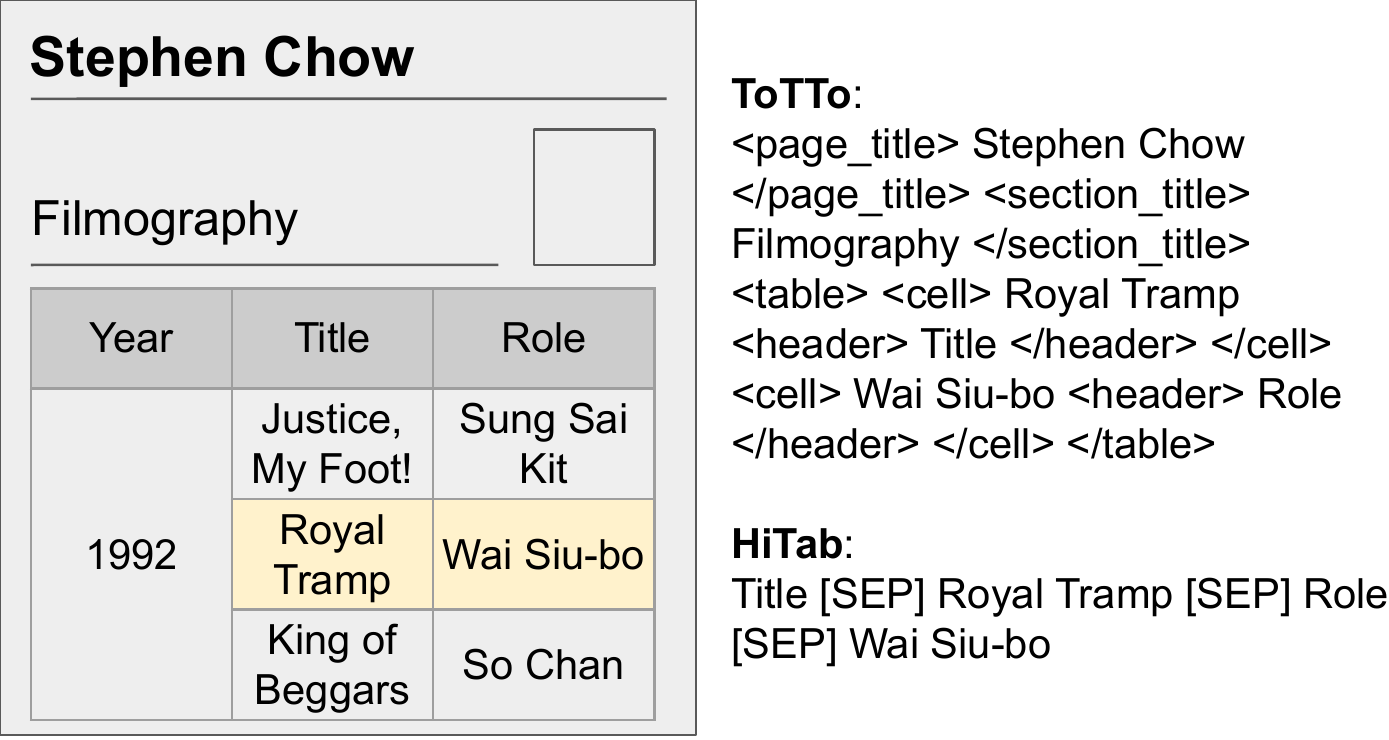}
  \end{center}
  \caption{Illustration of the input format for ToTTo and HiTab. Highlighted cells are marked in yellow (i.e. \textit{Royal Tramp} and \textit{Wai Siu-bo}).
    }
  \label{fig/input}
\end{figure}

\begin{table*}[t]
    \small
	\centering  %
	\setlength{\tabcolsep}{5pt}
	\renewcommand{\arraystretch}{1.1}
	\begin{tabular}{cccccccccc}
		\hlinewd{0.75pt}
		{\multirow{2}{*}{\textbf{Model}}}&\multicolumn{3}{c}{\textbf{Overall}}&\multicolumn{3}{c}{\textbf{Overlap}}&\multicolumn{3}{c}{\textbf{Non-Overlap}}\\
		\cmidrule(lr){2-4}
		\cmidrule(lr){5-7}
		\cmidrule(lr){8-10}
		&BLEU&PARENT&BLEURT&BLEU&PARENT&BLEURT&BLEU&PARENT&BLEURT \\
		\hline
		Pointer-Generator&41.6&51.6&0.076&50.6&58.0&0.244&32.2&45.2&-0.092 \\
		BERT-to-BERT&44.0&52.6&0.121&52.7&58.4&0.259&35.1&46.8&-0.017 \\ \hline
		T5-small & 45.3 & 57.0 & 0.187 & 52.7 & 61.0 & 0.316 & 37.8 & 53.0 & 0.057 \\
		\model (T5-small) & \textbf{47.4} & \textbf{57.8} & \textbf{0.207} & \textbf{55.6} & \textbf{62.3} & \textbf{0.337} & \textbf{39.1} & \textbf{53.3} & \textbf{0.077} \\ \hline
        T5-base & 47.4 & 56.4 & 0.221 & 55.5 & 61.1 & 0.344 & 39.1 & 51.7 & 0.098 \\
		\model (T5-base) & \textbf{48.4} & \textbf{58.1} & \textbf{0.222} & \textbf{56.1} & \textbf{62.4} & \textbf{0.345} & \textbf{40.4} & \textbf{53.9} & \textbf{0.099} \\ 
		\hlinewd{0.75pt}
	\end{tabular}
    \caption{Results on the ToTTo test set. Best scores are in bold.}
	\label{tb/totto}
\end{table*}

\begin{table}
    \small
    \centering
    \renewcommand{\arraystretch}{1.1}
    \begin{tabular}{c c c}
    	\hlinewd{0.75pt}
            {\textbf{Model}} & {BLEU} & {PARENT} \\
        \hline
        {Pointer-Generator} & {5.8} & {8.8} \\
        {BERT-to-BERT} & {11.4} & {16.7} \\ \hline
        {T5-small}   & {14.2} & {22.0}  \\
        {\model (T5-small)}  & {\textbf{15.7}} & {\textbf{23.8}} \\ \hline
        {T5-base} & {14.7} & {21.9} \\  
        {\model (T5-base)} & {\textbf{16.3}} & {\textbf{22.7}} \\ 
        \hlinewd{0.75pt}
    \end{tabular}
    \caption{Results on the HiTab test set. %
    }
    \label{tb/hitab}
\end{table}

As shown in \Cref{fig/input}, we use different input formats for ToTTo and Hitab following prior works \citep{kale2020text,cheng2021hitab}, since the tables and annotations in these two datasets are different.
For ToTTo, we follow the linearization procedure by \citet{kale2020text}. 
Specifically, the textual sequence consists of the page title, section title, table headers and cells.
Each cell may be associated with multiple row and column headers. 
Special markers are used to denote the begin and end of each field. 
Different from \citet{kale2020text}, we use the same markers for row headers and column headers. 
For HiTab, we follow the linearization procedure of \citet{cheng2021hitab}. 
Specifically, the textual sequence consists of highlighted cells and headers, headers of highlighted cells, and cells belong to highlighted headers. 
A universal separator token \texttt{[SEP]} is used. 
While our model can achieve consistently the same performance with any ordering of inputs, we adopt the same lexicographic as the layout-agnostic input format (\Cref{sec/method/aug}) to avoid uncertainty due to truncation and special markers.

\begin{table*}[t] 
    \small
	\centering  %
	\renewcommand{\arraystretch}{1.1}
	\setlength{\tabcolsep}{7pt}
	\begin{tabular}{lccccccccc}
		\hlinewd{0.75pt}
		{\multirow{2}{*}{\textbf{Model}}}&\multicolumn{3}{c}{\textbf{Overall}}&\multicolumn{3}{c}{\textbf{Overlap}}&\multicolumn{3}{c}{\textbf{Non-Overlap}}\\
		\cmidrule(lr){2-4}
		\cmidrule(lr){5-7}
		\cmidrule(lr){8-10}
        &Origin&Transform&$\Delta$&Origin&Transform&$\Delta$&Origin&Transform&$\Delta$\\
		\hline
		T5-small & 45.7 & 42.3 & -3.4 & 53.7 & 49.3 & -4.4 & 37.7 & 35.4 & -2.3 \\
		+ layout-agnostic input & 44.2 & 44.2 & \textbf{0} & 51.6 & 51.6 & \textbf{0} & 37.0 & 37.0 & \textbf{0} \\
		+ data augmentation & 45.3 & 44.4 & -0.9 & 52.8 & 52.0 & -0.8 & 37.9 & 37.0 & -0.9 \\
		\model (T5-small) & \textbf{47.5} & \textbf{47.5} & \textbf{0} & \textbf{55.5} & \textbf{55.5} & \textbf{0} & \textbf{39.5} & \textbf{39.5} & \textbf{0} \\ \hline
        T5-base & 47.4 & 42.9 & -4.5 & 55.8 & 50.7 & -5.1 & 39.2  & 35.4 & -3.8 \\
        + layout-agnostic input  & 46.2 & 46.2 & \textbf{0} & 54.3 & 54.3 & \textbf{0} & 38.3 & 38.3 & \textbf{0} \\
		+ data augmentation & 47.2 & 46.9 & -0.3 & 55.3 & 54.8 & -0.5 & 39.2 & 38.9 & -0.3 \\
		\model (T5-base) & \textbf{48.6} & \textbf{48.6} & \textbf{0} & \textbf{56.6} & \textbf{56.6} & \textbf{0} & \textbf{40.8} & \textbf{40.8} & \textbf{0} \\ 
		\hlinewd{0.75pt}
	\end{tabular}
    \caption{Robustness evaluation on ToTTo dev set. \textit{Origin} is the BLEU score on original tables, while \textit{Transform} is the BLEU score on transformed tables. All transformed tables are transposed, row shuffled and column shuffled. $\Delta$ is the difference between the two scores. Best scores in each group are in bold.}
	\label{tb/robust}
\end{table*}

\subsection{Main Results}
\label{sec/exp/main}
\Cref{tb/totto} shows model performance on ToTTo test set. 
Among the baselines, methods based on pretrained Transformer models (i.e. BERT-to-BERT and T5) outperform the others and T5 models perform the best.
Our method \model can be plugged into such models.
We compare our method with pure T5 models of different sizes, and \model consistently performs better.
Overall, \model (T5-small) achieves improvements of 2.1 BLEU points and 0.8 PARENT points in comparison with T5-small, and \model (T5-base) achieves improvements of 1.0 BLEU points and 1.7 PARENT points in comparison with T5-base.
These results indicate the importance of structure information, which is almost totally abandoned by baselines.
Further, the performance gain on tables both seen and unseen during training are significant.
Specifically, on the overlap subset, \model (T5-small) achieves improvements of 2.9 BLEU points and 1.3 PARENT points, and \model (T5-base) achieves improvements of 0.9 BLEU points and 1.3 PARENT points, indicating better intrinsic performance. 
On the non-overlap subset, \model (T5-small) achieves improvements of 1.3 BLEU points and 1.0 PARENT points, and \model (T5-base) achieves improvements of 1.3 BLEU points and 2.2 PARENT points, indicating \model is more generalizable to unseen tables.
We also observe that the improvement on BLEURT is not as much as the other two metrics.
It is reasonable as BLEURT is trained with machine translation annotations and synthetic data by mask filling, backtranslation and word drop.
These training data ensures its robustness to surface generation but not reasoning-based generation.
Although the effectiveness of BLEURT is verified on an RDF-to-text dataset, tabular data holds different properties with RDF data\footnote{For example, in the ToTTo dataset, 21\% samples requires reasoning while 13\% samples requires comparison.}.

Results on HiTab in \Cref{tb/hitab} further verify the effectiveness and generalizability of \model.
For different model sizes, \model consistently performs better than T5 models.
We also observe that on this dataset the model with highest BLEU score is not the model with highest PARENT score.
It is partially because of the annotations. 
Many numbers appear in both tables and target sentences are of different precision.
Copying such numbers from tables to generated sentences may increase PARENT score but reduce BLEU score.

\subsection{Robustness Evaluation} 
\label{sec/exp/robust}
To further evaluate model robustness against content-invariant perturbations on tables, we create a harder version of the ToTTo dev set, where each table is perturbed with a combination of row-wise shuffling, column-wise shuffling and table transpose.
Especially, models can no longer benefit from memorizing the layout of tables appearing in both the training set and the dev set.
We compare four methods based on T5, including the basic version proposed by \citet{kale2020text}, enhanced T5 with the layout-agnostic input or data augmentation (\Cref{sec/method/aug}), and T5 incorporated in \model.

According to the results shown in
\Cref{tb/robust},
vanilla T5 models face a severe performance drop when content-invariant perturbations are introduced.
Overall, BLEU scores drop by 3.4 for T5-small, and 4.5 for T5-base.
We also observe that the performance drop on overlap subset is larger than on non-overlap subset.
This indicates that the performance gain of T5 models is somehow due to their memory of some tables existing in the training set, which is however brittle and not generalizable.
Applying layout-agnostic input format, which linearizes tables by lexicographic order instead of cell index order, ensures models to return stable predictions, but results in worse overall performance due to the loss of structural information.
Not surprisingly, layout-agnostic input causes performance drops by 1.5 BLEU points and 1.2 BLEU points to T5-small and T5-base on original dev set.

Another common way to improve model robustness %
is to increase the diversity of training instances with data augmentation.
We augment the original training set by 8-fold using the three content-invariant transformation operations and their combinations. %
Training with augmented data reduces the gap between model performance on original tables and transformed tables.
However, data augmentation is never exhaustive enough to guarantee true equivariance.
Also, this introduces different variants of the same table into the training set, so there is a gap between the same table in training set and dev set. 
As a result, the performance on overlap subset is slightly worse than without data augmentation, but the performance on non-overlap subset is not negatively influenced.
\model guarantees consistent predictions towards content-invariant table transformations while achieving the best performance.
In comparison with using layout-agnostic input format which also guarantees equivariance, \model (T5-small) provides additional 3.3 BLEU points, and \model (T5-base) provides additional 2.4 BLEU points on original dev set.

\begin{table}[t]
    \small
	\centering  %
	\renewcommand{\arraystretch}{1.1}
	\setlength{\tabcolsep}{7pt}
	\begin{tabular}{ccccc}
		\hlinewd{0.75pt}
		\textbf{Att}&\textbf{Pos}&Overall&Overlap&Non-Overlap\\
		\hline
		 - & - & 45.7 &  53.7 & 37.7  \\ \hline
		\checkmark & - & 47.0	& 54.4	&	39.6   \\
		\checkmark & \checkmark & 47.5	& 55.5	& 39.5  \\
		\hlinewd{0.75pt}
	\end{tabular}
    \caption{Ablation study on ToTTo dev set. Scores are BLEU. \textit{Att} and \textit{Pos} denote structure-aware self-attention and transformation-invariant positional encoding.}
	\label{table/ablation}
\end{table}

\subsection{Ablation Study}
\label{sec/exp/ablation}
To help understand the effect of two key mechanisms in transformation-invariant graph masking, we hereby present %
ablation study results in \Cref{table/ablation}.

\stitle{Structure-Aware Self-Attention}
We examine the effectiveness of structure-aware self-attention.
In comparison with original (fully-connected) self-attention, incorporating structural information by pruning attention flows can improve the overall performance by 1.3 BLEU points.
Detailed scores on two subsets show that both tables seen and unseen during training can benefit from structural information.
The consistent improvements on two subsets indicate that structure-aware self-attention improves model ability of capturing cell relations rather than memorizing tables.

\stitle{Transformation-Invariant Positional Encoding}
We further test the effectiveness of transformation-invariant positional encoding.
We observe that although this technique is mainly designed for ensuring model robustness towards layout changes, it can bring an additional improvement of 0.5 BLEU points to overall performance.
Interestingly, the improvement is mainly on the overlap subset.
We attribute it to the fact that the same table in training and dev sets may have different highlighted cells, so that memorizing the layout information in the training set hinders in-domain generalization.

\section{Related Work}

We review two relevant research topics. Since both topics have a large body of work, we provide a selected summary.

\stitle{Table-to-text Generation}
Table-to-text generation seeks to generate textual descriptions for tabular data. In comparison to text-to-text generation, the input of table-to-text generation is semi-structured data.
Early studies adapt the encoder-decoder framework to data-to-text generation with encoders aggregating cell information \citep{lebret2016neural, wiseman2017challenges, bao2018table}.
Followed by the success of massively pre-trained sequence-to-sequence Transformer models \citep{raffel2020exploring, lewis2020bart}, recent SOTA systems apply these models to table-to-text generation \citep{kale2020text, su2021plan}, where the input table is linearized to a textual sequence.

A table can include ample information and it is not always able to be summarized in one sentence.
A line of work learns to %
generate selective descriptions by paying attention to key information in the table \citep{perez2018bootstrapping, ma2019key}.
However, multiple statements can be entailed from a table when %
different parts of the table are focused on.
To bridge this gap, \citet{parikh2020totto} proposes controlled table-to-text generation, %
allowing the generation process to react differently according to distinct highlighted cells.
As highlighted cells can be at any positions and of arbitrary numbers, simple linearization, which breaks the layout structure, hinders relations among cells from being captured, therefore %
causing unreliable or hallucinated descriptions to be generated. 

A few prior studies introduce structural information to improve model performance on table-to-text generation, either by incorporating token position \citep{liu2018table}, or by aggregating row and column level information \citep{bao2018table, nema2018generating, jain2018mixed}.
However, none of existing methods can be directly applied to pretrained Transformer-based generative models, especially when we want to ensure model robustness to content-invariant table transformations.
Our method enforces both structure-awareness and transformation-invariance to such models.

\stitle{Equivariant Representation Learning}
Equivariance is a type of prior knowledge existing broadly in real-world tasks.
Earlier studies show that incorporating equivariance learning can improve visual perception model
robustness against turbulence caused by geometric transformations, 
such as realizing translation, rotation, and scale equivariance of images \citep{lenc2015understanding, worrall2017harmonic, ravanbakhsh2017equivariance, sosnovik2019scale, yang2020rotation}.
The input to those tasks presents unstructured information, 
and several geometrically invariable operations are incorporated in neural networks to realize the aforementioned equivariance properties.
For example, Convolutional Neural Networks (CNNs) are equivariant to translations in nature \citep{lenc2015understanding}.
Harmonic Networks and Spherical CNNs extend the equivariance of CNNs to rotations \citep{worrall2017harmonic, esteves2018learning}.
Group Equivariant Convolutional Networks are equivariant to more spatial transformations including translations, rotations and reflections \citep{cohen2016group}.
Nonetheless, none of these geometrically invariable techniques can be directly applied to Transformer-based generative models to ensure equivariance on (a part of) structured tabular data, which is exactly the focus of this work. 
Our method realizes equivariant intermediate representations against content-invariant table transformations in table-to-text generation.

Some other works, while not explicitly using equivariant model structures, seek to realize equivariant representations by augmenting more diverse changes into training data \citep{chen2020group, wu2020generalization}. 
Although the model can benefit from %
seeing more diverse inputs involving content-invariant transformations \citep{wu2020generalization}, this strategy has two drawbacks. %
Specifically, the augmented data, while introducing much computational overhead to training, are never exhaustive enough to guarantee true equivariance.
By contrast, our method guarantees equivariance through the neural network design and do not introduce any training overhead.

\section{Conclusion}

We propose \model, a structure-aware equivariance learning framework for controlled table-to-text generation.
Our experimental results verify the importance of structure-awareness and transformation-invariance, two key properties enforced in \model, towards precise and robust description generation for tabular content.
The proposed properties and equivariance learning framework aligns well with the nature of information organized in tables. %
Future research can consider extending the structure-aware equivariance learning framework to other data-to-text generation tasks \citep{koncel2019text,nan2021dart}, tabular reasoning or retrieval tasks \cite{gupta2020infotabs,wang2021retrieving,wang2021table,eisenschlos2021mate}, and pretraining representation on textual and tabular data \cite{yin2020tabert,herzig2020tapas,iida2021tabbie}.

\section*{Acknowledgement}

We appreciate the reviewers for their insightful comments and suggestions.
This work is partly supported by the National Science Foundation of United States Grant IIS 2105329,
and partly by the Air Force Research Laboratory under agreement number FA8750-20-2-10002.

\section*{Ethical Considerations}

This work seeks to develop a structure-aware equivariance learning framework for table-to-text generation. 
Since the proposed method focuses on improving prior generation systems by better utilization of structural information, it does not introduce bias towards specific content.
The distinction between beneficial use and harmful use depends mainly on the data.
Proper use of the technology requires that input corpora are legally and ethically obtained.
We conduct experiments on two open benchmark in the way they intended to.
Although we create a harder version of ToTTo dev set, the table transformation operations we use are content-invariant, whereas the ground-truth generation remains the same as it is in the original dataset, ensuring no further social bias is introduced.

\bibliography{ref}
\bibliographystyle{format/acl_natbib}

\end{document}